\documentclass{article}

    \PassOptionsToPackage{numbers, compress}{natbib}

\usepackage[preprint]{neurips_2019}

\usepackage[utf8]{inputenc} 
\usepackage[T1]{fontenc}    
\usepackage{hyperref}       
\usepackage{url}            
\usepackage{booktabs}       
\usepackage{amsfonts}       
\usepackage{nicefrac}       
\usepackage{microtype}      

\usepackage{algorithm}
\usepackage{algorithmic}

\usepackage{amsmath}
\usepackage{bm}
\usepackage{graphicx}
\usepackage{wrapfig}
\usepackage{xcolor}

\title{Defending Adversarial Attacks by Correcting {\em logits}}

\author{%
Yifeng Li\textsuperscript{1}, Lingxi Xie\textsuperscript{2}, Ya Zhang\textsuperscript{1}, Rui Zhang\textsuperscript{1}, Yanfeng Wang\textsuperscript{1}, Qi Tian\textsuperscript{2} \\
\textsuperscript{1}Cooperative Medianet Innovation Center, Shanghai Jiao Tong University \\
\textsuperscript{2}Noah's Ark Lab, Huawei Inc. \\
\small{\texttt{tobylyf@live.com}\quad\texttt{198808xc@gmail.com}} \\
\small{\texttt{\{ya\_zhang,zhang\_rui,wangyanfeng\}@sjtu.edu.cn}\quad\texttt{tian.qi1@huawei.com}} \\
}

\begin{document}

\maketitle

\begin{abstract}
Generating and eliminating adversarial examples has been an intriguing topic in the field of deep learning. While previous research verified that adversarial attacks are often fragile and can be defended via image-level processing, it remains unclear how high-level features are perturbed by such attacks. We investigate this issue from a new perspective, which purely relies on {\em logits}, the class scores before softmax, to detect and defend adversarial attacks. Our {\em defender} is a two-layer network trained on a mixed set of clean and perturbed {\em logits}, with the goal being recovering the original prediction. Upon a wide range of adversarial attacks, our simple approach shows promising results with relatively high accuracy in defense, and the defender can transfer across attackers with similar properties. More importantly, our defender can work in the scenarios that image data are unavailable, and enjoys high interpretability especially at the semantic level.
\end{abstract}

\section{Introduction}
\label{introduction}

In the past years, with the blooming development of deep learning, researchers started focusing on the `dark side' of deep neural networks, such as the lack of interpretability. In this research field, an important topic is the existence of adversarial examples, which claims that slightly modifying the input image, sometimes being imperceptible to human, can lead to a catastrophic change in the output prediction~\citep{Szegedy2014IntriguingPO,Goodfellow2015ExplainingAH}. Driven by the requirements of understanding deep networks and securing AI-based systems, there emerge a lot of efforts in generating adversarial examples to attack deep networks~\citep{Szegedy2014IntriguingPO,Goodfellow2015ExplainingAH,MoosaviDezfooli2016DeepFoolAS,Carlini2017TowardsET,Kurakin2017AdversarialEI,Dong2018BoostingAA} and, in the opposite direction, designing algorithms to recognize adversarial examples and thus eliminate their impacts~\citep{Papernot2016DistillationAA,Kurakin2017AdversarialML,Madry2018TowardsDL,Tramr2018EnsembleAT,Guo2018CounteringAI,Xie2018MitigatingAE,Prakash2018DeflectingAA,Liao2018DefenseAA}.

This paper focuses on the defense part. On the one hand, researchers have demonstrated that adversarial attacks are fragile and thus their impacts can be weakened or eliminated by some pre-processing on the input image~\citep{Guo2018CounteringAI,Xie2018MitigatingAE,Prakash2018DeflectingAA,Liao2018DefenseAA}; on the other hand, there are also efforts in revealing how such attacks change mid-level and high-level features and thus eventually break up prediction~\citep{Liao2018DefenseAA,Xie2018FeatureDF}. Here we make a hypothesis: the cues of recovering low-level features and high-level features from adversarial attacks are quite different: for low-level features, it is natural to use image-space priors such as intensity distribution and smoothness to filter out adversaries; for high-level features, however, the cues may lie in the intrinsic connections between the semantics of different classes.

To the best of our knowledge, this is the first work that purely focuses on defending adversarial attacks from topmost semantics --- this forces our system to find semantic cues, which were not studied before, for defending adversaries. To this end, we study {\em logits}, which are the class scores produced by the last fully-connected layers --- before being fed into softmax normalization. {\bf The major discovery of this paper lies in that we can take merely {\em logits} as input to defend adversarial attacks.} To reveal this, we first train a $1\rm{,}000$-way classifier on the ILSVRC2012 dataset~\cite{Russakovsky2015ImageNetLS}. Then, we use an off-the-shelf attacker such as PGD~\citep{Madry2018TowardsDL} to generate adversarial examples, in particular {\em logits} are recorded. Finally, we train a two-layer fully-connected network to predict the original label. We mix both clean and adversarial {\em logits} into the training set, so that the trained defender has the ability of recovering attacked prediction meanwhile keeping non-attacked prediction unchanged.

Working at the semantic level, our approach enjoys two-way benefits. First, we only use {\em logits} for defense, which reduces the requirements of image data and mid-level features --- this is especially useful in the scenarios that image data are inaccessible due to security reasons, and that very low bits of data can be transferred for defense (note that compared to low-level and mid-level features, {\em logits} have a much lower dimension). Second and more importantly, we make it possible to explain how adversaries are defended by directly digging into the two-layer network. An interesting phenomenon is that almost all attackers can leave fingerprints on a few fixed classes, which are named {\em supporting classes}, even if the adversarial class can vary among the entire dataset. Our defender works by detecting these supporting classes, and whether a defender can transfer to different attacks and/or network models can be judged by evaluating how these supporting classes overlap. By revealing these properties, we move one step further towards understanding the mechanisms of adversaries.



\section{Backgrounds}
\label{backgrounds}

In this section, we provide background knowledge about adversarial examples and review some representative attack and defense methods. Let $\mathbf{x}$ denote a clean/natural image sample, and $y$ is the corresponding label which has $C$ choices. A deep neural network $f(\cdot)$ is defined as $y = f(\mathbf{x}): \mathcal{X} \to \mathcal{Y}$. In practice, given an input $\mathbf{x}$, the classifier $f(\cdot)$ outputs a final feature vector $\mathbf{z} \doteq Z(\mathbf{x})$ called {\em logits} where each element $z_k$ corresponds to the $k$-th class. The {\em logits} are then fed into a softmax function to produce the predicted probability $p(y|\mathbf{x}) = \mathrm{softmax}(\mathbf{z})$, and the predicted class is chosen by $f(\mathbf{x}) = {\arg \max}_y p(y|\mathbf{x})$. There exist adversarial attacks that went beyond classification~\citep{Xie2017AdversarialEF}, but we focus on classification because it is the fundamental technique extended to other tasks.

\subsection{Adversarial attacks}
\label{backgrounds:attacks}

An adversarial example $\mathbf{x}^\ast$ is crafted by adding an imperceptible perturbation to input $\mathbf{x}$, so that the prediction of classifier $f(\cdot)$ becomes incorrect. If an attack aims at just forcing the classifier to produce a wrong prediction, {\em i.e.}, $f(\mathbf{x}^\ast) \neq f(\mathbf{x})$, it is called a {\em non-targeted} attack. On the other hand, a {\em targeted} attack is designed to cause the classifier to output a specific prediction as $f(\mathbf{x}^\ast) = y^\ast$, where $y^\ast$ is the target label and $y^\ast \neq f(\mathbf{x})$. In this paper, we mainly focus on the non-targeted attack, as it is more common and easier to achieved by a few recent approaches~\citep{Goodfellow2015ExplainingAH,Kurakin2017AdversarialEI,Madry2018TowardsDL,Dong2018BoostingAA,MoosaviDezfooli2016DeepFoolAS,Carlini2017TowardsET}. Another way of categorizing adversarial attacks is to consider the amount of information the attacker has. A {\em white-box} attack means that the adversary has full access to both the target model $f(\cdot)$ and how the defender works. In the opposite, a {\em black-box} attack indicates that the attacker knows neither the classifier nor the defender. Also there are some intermediate cases, termed the {\em gray-box} attack, in which partial information is unknown. Generally, two types of methods to generate adversarial examples were most frequently used by researchers:

{\bf Gradient-sign ($\bm{\ell_{\infty}}$) methods.}\quad To generate adversarial examples efficiently, \citet{Goodfellow2015ExplainingAH} proposed the Fast Gradient Sign Method (FGSM) that took a single step in the signed-gradient direction of the cross-entropy loss, based on an assumption that the classifier is approximately linear locally. However, this assumption may not hold perfectly and the success rates of FGSM are relatively low. To address this issue, \citet{Kurakin2017AdversarialEI} proposed the Basic Iterative Method (BIM) that applied fast gradient iteratively with smaller steps. \citet{Madry2018TowardsDL} further extended this approach to a `universal' first-order adversary by introducing a random start. The proposed Projected Gradient Descent (PGD) method served as a very strong $\ell_\infty$ attack in the white-box scenario. \citet{Dong2018BoostingAA} stated that FGSM adversarial examples `under-fit' the target model and BIM `over-fit' it, respectively, making them hard to transfer across models. They proposed the Momentum-based Iterative Method (MIM) that integrated a momentum term into the iterative process to stabilize update directions.

{\bf $\bm{\ell_2}$ attacks.}\quad \citet{MoosaviDezfooli2016DeepFoolAS} also considered a linear approximation of the decision boundaries and proposed the DeepFool attack to generate adversarial examples that minimize perturbations measured by $\ell_2$-norm. It iteratively moved an image towards the nearest decision boundary until the image crosses it and becomes misclassified. Unlike previous iterative gradient-based approaches, \citet{Carlini2017TowardsET} proposed to directly minimize a loss function so that the generated perturbation makes the example adversarial meanwhile its $\ell_2$-norm is optimized to be small. The C\&W attack always generates very strong adversarial examples with low distortions.

\subsection{Adversarial defenses}
\label{backgrounds:defenses}

Many methods defending against adversarial examples have been proposed recently. These methods can be roughly divided into two categories. One type of defenses worked by modifying the training or inference strategies of models to improve their inherent robustness against adversarial examples. Adversarial training is one of the most popular and effective method of this kind~\citep{Goodfellow2015ExplainingAH,Kurakin2017AdversarialML,Madry2018TowardsDL,Tramr2018EnsembleAT,Kannan2018AdversarialLP}. It augmented or replaced the training data with adversarial examples and aimed at training a robust model being aware of the existence of adversaries. While effectively improving the robustness to seen attacks, this type of approaches often consumed much more computational resources than normal training. Other methods include defensive distillation~\citep{Papernot2016DistillationAA}, saturating networks~\citep{Nayebi2017BiologicallyIP}, thermometer encoding~\citep{Buckman2018ThermometerEO}, {\em etc.}, which benefited from gradient masking effect~\citep{Papernot2017PracticalBA} or obfuscated gradients~\citep{Athalye2018ObfuscatedGG} but were still vulnerable to black-box attacks.

Another line of defenses was based on removing adversarial perturbations by processing the input image before feeding them to the target model. \citet{Dziugaite2016ASO} studied JPEG compression to reduce the effect of adversarial noises. \citet{Osadchy2017NoBE} applied a set of filters like the median filter and averaging filter to remove perturbations. \citet{Guo2018CounteringAI} studied five image transformations and showed that total-variance minimization and image quilting were effective for defense. \citet{Xie2018MitigatingAE} leveraged random resizing and padding to mitigate adversarial effects. \citet{Prakash2018DeflectingAA} proposed pixel deflection that redistributed pixel values locally and thus broke adversarial patterns. \citet{Liao2018DefenseAA} utilized  a high-level representation guided denoiser to defend adversaries. Nonetheless, these methods did not show high effectiveness against very strong perturbations.

Another approach most similar to our work is \citep{Roth2019TheOA}, which detected and attempted to recover adversarial examples by measuring how {\em logits} change when the input is randomly perturbed. Our method differs from it in that we purely leverage the final {\em logits} but no input image.

\section{Settings: dataset, network models and attacks}
\label{settings}

Throughout this paper, we evaluate our approach on the ILSVRC2012 dataset~\citep{Russakovsky2015ImageNetLS}, a large-scale image classification task with $C=1\rm{,}000$ classes, around $1.3\mathrm{M}$ training images and $50\mathrm{K}$ validation images. This dataset was also widely used by other researchers for adversarial attacks and defenses. We directly use a few pre-trained network models from the model zoo of the PyTorch platform~\citep{Paszke2017AutomaticDI}, including VGG-16~\citep{Simonyan2015VeryDC}, ResNet-50~\citep{He2016DeepRL} and DenseNet-121~\citep{Huang2017DenselyCC}, which report top-1 classification accuracies of $73.48\%$, $76.13\%$ and $74.47\%$, respectively.

As for adversarial attacks, we evaluate a few popular attacks mentioned in Section~\ref{backgrounds:attacks} using the CleverHans library~\citep{papernot2016TechnicalRO}, including PGD~\citep{Madry2018TowardsDL}, MIM~\citep{Dong2018BoostingAA}, DeepFool~\citep{MoosaviDezfooli2016DeepFoolAS} and C\&W~\citep{Carlini2017TowardsET}, each of which causes dramatic accuracy drop to the pre-trained models (see Section~\ref{approach:results}). Below we elaborate the technical details of these attacks, most of which simply follow their original implementations.

{\bf PGD}~\citep{Madry2018TowardsDL} adversarial examples are generated by the following method:
\begin{equation}
    \mathbf{x}_0^\ast \sim \mathcal{B}_{\epsilon}^{\infty}(\mathbf{x}),\quad
    \mathbf{x}_{t+1}^\ast = \Pi_{\mathcal{B}_{\epsilon}^{\infty}(\mathbf{x})} \left( \mathbf{x}_t^\ast + \alpha \cdot \mathrm{sign}(\nabla_{\mathbf{x}_t^\ast}J(\mathbf{x}_t^\ast, y)) \right),\quad
    \mathbf{x}^\ast = \mathbf{x}_T^\ast.
\end{equation}
Here, $\mathcal{B}_{\epsilon}^{\infty}(\mathbf{x})$ is the $\ell_{\infty}$ ball around $\mathbf{x}$, $\mathbf{x}_0^\ast$ is randomly sampled from inside the ball and $\Pi_{\mathcal{B}_{\epsilon}^{\infty}(\mathbf{x})} (\cdot)$ means clipping the examples so that they stay in the ball and satisfy the $\ell_{\infty}$ constraint. We set the maximum perturbation size to be $\epsilon=\nicefrac{16}{255}$, number of iterations $T=10$ and step size $\alpha=\nicefrac{2}{255}$.

{\bf MIM}~\citep{Dong2018BoostingAA} adversarial examples are generated using the following algorithm:
\begin{equation}
\begin{split}
    \mathbf{g}_0 &= \mathbf{0},\quad
    \mathbf{g}_{t+1} = \mu \cdot \mathbf{g}_t + \nabla_{\mathbf{x}_t^\ast}J(\mathbf{x}_t^\ast, y)/||\nabla_{\mathbf{x}_t^\ast}J(\mathbf{x}_t^\ast, y)||_1, \\
    \mathbf{x}_0^\ast &= \mathbf{x},\quad
    \mathbf{x}_{t+1}^\ast = \mathbf{x}_t^\ast + \alpha \cdot \mathrm{sign}(\mathbf{g}_{t+1}),\quad
    \mathbf{x}^\ast = \mathbf{x}_T^\ast.
\end{split}    
\end{equation}
With maximum perturbation size $\epsilon=\nicefrac{16}{255}$, we set the number of iterations $T=10$, step size $\alpha=\nicefrac{\epsilon}{T}$ and decay factor $\mu=1.0$.

{\bf DeepFool}~\citep{MoosaviDezfooli2016DeepFoolAS} iteratively finds the minimal perturbations to cross the linearization of the nearest decision boundaries. $10$ most-likely classes are sampled when looking for the nearest boundaries, and the maximum number of iterations is set to be $100$.

{\bf C\&W}~\citep{Carlini2017TowardsET} jointly optimizes the norm of perturbation and a hinge loss of giving an incorrect prediction. For efficiency, we set the maximum number of iterations to $25$ with $4$ binary search steps for the trade-off constant. The initial trade-off constant, hinge margin and learning rate are set to be $10$, $0$ and $0.01$, respectively.

\section{Defending adversarial attacks by {\em logits}}
\label{approach}

Our goal is to detect and defend adversarial attacks by {\em logits}. The main reason of investigating {\em logits} lies in two parts. First, it makes the algorithm easier to be deployed in the scenarios that the original input image and/or mid-level features are not available. Second, it enables us to understand the mechanisms of adversaries as well as how they are related to the interpretability of deep networks.

Mathematically, we are given a set of {\em logits} $\mathbf{z}$ which is the final output of the deep network $f(\cdot)$ before softmax and the input is either a clean image $\mathbf{x}$ or an adversarial image $\mathbf{x}^\ast$. Most often, $f(\mathbf{x})$ leads to the correct class while $f(\mathbf{x}^\ast)$ can be dramatically wrong. The goal is to recover the original label from $f(\mathbf{x}^\ast)$ while remaining the prediction of $f(\mathbf{x})$ unchanged.


\subsection{The possibility of defending adversaries by {\em logits}}
\label{approach:possibility}

\begin{wrapfigure}{r}{6cm}
\centering
\includegraphics[width=\linewidth]{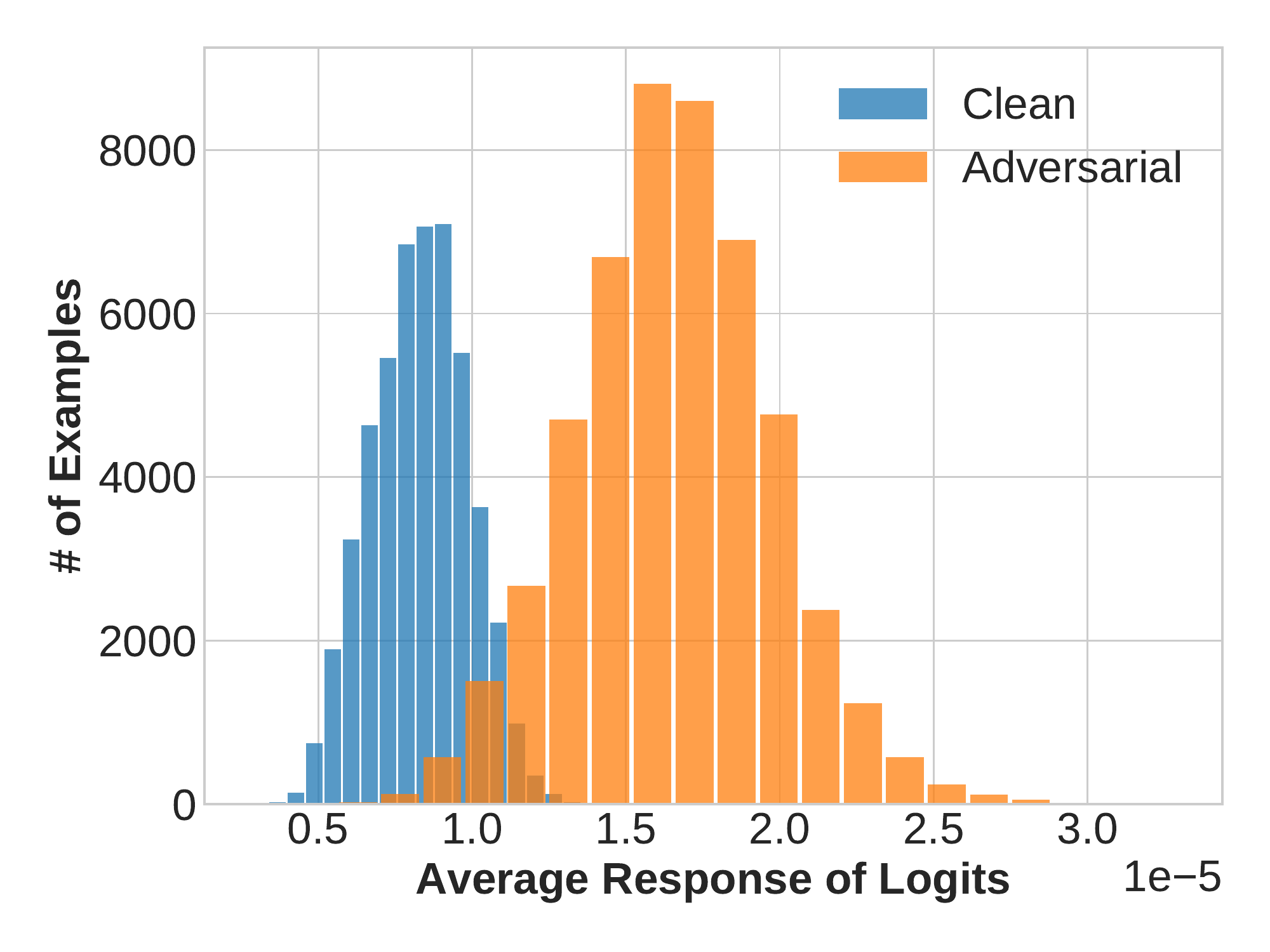}
\caption{
Average response of {\em logits} on clean and PGD adversarial examples, counted on the validation set of ILSVRC2012. We fix the number of bins to be $20$ for both types of data. In most cases, the PGD attack has made the mean value of {\em logits} greater.}
\label{fig:logits}
\end{wrapfigure}

The basis of our research lies in the possibility of defending adversarial attacks by merely checking the {\em logits}. In other words, the numerical values of {\em logits} before and after being attacked are quite different. To reveal this, we use an example of the PGD attack~\citep{Madry2018TowardsDL}, while we also observe similar phenomena in other cases. We apply PGD over all $50\rm{,}000$ validation images, and plot the distributions of the average values of the {\em logits}, before and after the dataset is attacked. Figure~\ref{fig:logits} shows the histogram of the average response. One can observe that adversaries cast significant changes which are obvious even under such simple statistic. 

The above experiment indicate that adversarial attacks indeed leave some kind of `fingerprints' in high-level feature vectors, in particular {\em logits}. This is interesting because {\em logits} features are produced by the last layer of a deep network, so (i) do not contain any spatial information and (ii) each element in it corresponds to one class of the dataset. The former property makes our defender quite different from those working on image data or intermediate features~\citep{Dziugaite2016ASO,Osadchy2017NoBE,Guo2018CounteringAI,Xie2018MitigatingAE,Prakash2018DeflectingAA,Xie2018FeatureDF}, which often made use of spatial information for noise removal. The latter property eases the defender to learn inter-class relationship to determine whether a case was adversarial and to recover it from attacks. However, due to the high dimensionality of {\em logits} --- the dimension is $C=1\rm{,}000$ in our problem and can be higher in the future, it is difficult to achieve the goal of defense upon a few fixed rules. This motivates us to design a learning-based approach for this purpose.

\subsection{Adversarial {\em logits} correction}
\label{approach:correction}

We first consider training a defender for a specific deep network $f(\cdot)$, say ResNet-50~\citep{He2016DeepRL}, and a specific attack $A(\cdot)$, say PGD~\citep{Madry2018TowardsDL}. We should discuss on the possibility of transfer this defender to other combinations of network and attack in Section~\ref{explanation}, after we explain how it works. The first step is to build up a training dataset. Recall that our goal is to recover the prediction of contaminated data while remain that of clean data unchanged, so we collect both types of data by feeding each training case $\mathbf{x}$ into $A(\cdot)$ so as to produce its adversarial version $\mathbf{x}^\ast=A(\mathbf{x})$. Then, both $\mathbf{x}$ and $\mathbf{x}^\ast$ are fed into $f(\cdot)$ and the corresponding {\em logits} are recorded as $\mathbf{z}$ and $\mathbf{z}^\ast$, respectively. We ignore the softmax layer to generate the final labels $y$ and $y^\ast$, since it is trivial and does not impact the defense process.

We then train a two-layer fully-connected network upon these $C$-dimensional {\em logits}. Specifically, the output layer is still a $C$-dimensional vector which represents the corrected {\em logits}, and the hidden layer contains $D=10\rm{,}000$ neurons, each of which is equipped with a ReLU activation function~\citep{Krizhevsky2012ImageNetCW} and Dropout~\citep{Hinton2012ImprovingNN} with a keep ratio of $0.5$. This design is to maximally simplify the network structure to facilitate explanation, while being able to deal with non-linearity in learning relationship between classes. The relatively large amount of hidden neurons eases learning complicated rules for {\em logits} correction. While we believe more complicated network design can improve its ability of defense, this is not the most important topic of this paper. As a side note, we tried to simply use more hidden layers to train a deeper correction network, but observed performance decrease on adversarial examples. The training process of this defender network follows a standard gradient-based optimization, with very few bells and whistles added. A detailed illustration is provided in Algorithm~\ref{alg:correction}. Note that we mix clean and contaminated images with a probability $p$ that a clean image is sampled. This strategy is simple yet effective in enabling the defender to maintain the original prediction of clean data.


\begin{algorithm}[t]
\caption{Adversarial {\em logits} Correction}
\label{alg:correction}
\begin{algorithmic}[1]
\REQUIRE Clean training set $\mathcal{S} = \{\mathbf{x}_n, y_n\}_{n=1}^N$, pre-trained classifier $f(\cdot)$, adversarial attacker $A(\cdot)$;
\REQUIRE \# of hidden neurons $D$, \# of training iterations $T$, clean training probability $p$;
\ENSURE {\em logits} correction network $g(\cdot)$;
\STATE Perturb $\mathcal{S}$ with $A(\cdot)$ and obtain an adversarial counterpart $\mathbf{x}_n^\ast$ for each $\mathbf{x}_n$;
\STATE Feed all examples, $\mathbf{x}_n$ and $\mathbf{x}_n^\ast$, into $f(\cdot)$ and extract the corresponding {\em logits} $\mathbf{z}_n$ and $\mathbf{z}_n^\ast$;
\STATE Randomly initialize $g(\cdot)$ as a two-layer fully-connected network with $D$ hidden neurons;
\FOR{$t \gets 1$ to $T$}
    \STATE Sample a mini-batch $\mathcal{B}_t = \{\mathbf{z}_m^\circ, y_m\}_{m=1}^M$ from $\mathcal{S}$, in which $\mathbf{z}_m^\circ$ takes either $\mathbf{z}_m$ or $\mathbf{z}_m^\ast$, with a probability of $p$ to take the clean {\em logits}, $\mathbf{z}_m$;
    \STATE Do the training step of $g(\cdot)$ using $\mathcal{B}_t$ and the cross-entropy loss;
\ENDFOR
\RETURN $g(\cdot)$.
\end{algorithmic}
\end{algorithm}

Before going into experiments, we briefly discuss the relationship between our approach, adversarial {\em logits} correction, and prior work. The {\bf first} family was named `adversarial training'~\citep{Goodfellow2015ExplainingAH,Kurakin2017AdversarialML,Madry2018TowardsDL,Tramr2018EnsembleAT,Kannan2018AdversarialLP}, which generated adversarial examples online and added them into the training set. Our approach is different in many aspects, including the goal (we hope to defend adversaries towards trained models rather than increasing the robustness of models themselves) and the overheads (we do not require a costly online generation process). More importantly, we generate and defend adversaries in the level of {\em logits}, which, to the best of our knowledge, was never studied in the previous literature. The {\bf second} one was often called `learning from noisy labels'~\citep{Angluin1987LearningFN}, in which researchers proposed to estimate a transition matrix~\citep{Goldberger2017TrainingDN,Patrini2017MakingDN} in order to address the corrupted labels. Although our approach also relies on the same assumption that corrupted or adversarial labels can be recovered by checking class-level relationship, we emphasize that the knowledge required for correcting adversaries is quite different from that for correcting label noises. This is because the effects brought by adversaries are often less deterministic, {\em i.e.}, the adversarial label is often completely irrelevant to the original one, but a noisy label can somewhat reflect the correct one. This is also the reason why we used a learning-based approach and built a two-layer network with a large number of hidden neurons. In practice, using a single-layer network or reducing hidden neurons can cause dramatic accuracy drop in recovering adversaries, {\em e.g.}, under the PGD attack towards ResNet-50, the recovery rate of a two-layer network is $75.4\%$ but drops to $41.9\%$ when a single-layer network is used (see the next part for detailed experimental settings).

\subsection{Experimental results}
\label{approach:results}

We first evaluate our approach on the PGD attack~\citep{Madry2018TowardsDL} with different pre-trained network models. We use the Adam optimizer~\citep{Kingma2015AdamAM} to train the defenders for $50$ epochs on the entire ILSVRC2012~\citep{Russakovsky2015ImageNetLS} training set, with a batch size of $256$, learning rate of $0.00005$, weight decay of $0.0001$, and a probability of choosing clean data, $p$, of $0.3$. We evaluate both a {\em full} testing set ($50\rm{,}000$ images) and a {\em selected} testing set ($1\rm{,}000$ correctly classified images, to compare with other attackers). Table~\ref{tab:network} shows the classification accuracy of different networks on both clean and PGD-attacked examples with and without adversarial {\em logits} correction. One the one hand, although the PGD attack is very strong in this scenario, reducing the classification accuracy of all networks to nearly $0\%$, it is possible to recover the correct prediction by merely checking {\em logits}, which (i) do not preserve any spatial information, and (ii) as high-level features, are supposed to be perturbed much more severely than the input data~\citep{Liao2018DefenseAA,Xie2018FeatureDF}. After correction, the classification accuracy on VGG-16~\citep{Simonyan2015VeryDC} and ResNet-50~\citep{He2016DeepRL} are only reduced by $6.01\%$ and $3.74\%$, respectively. To the best of our knowledge, this is the only work which purely relies on {\em logits} to defend adversaries, so it is difficult to compare these results with prior work. On the other hand, on clean (un-attacked) examples, our approach reports slight accuracy drop, because some of them are considered to be adversaries and thus mistakenly `corrected'. This is somewhat inevitable, because {\em logits} lose all spatial information and non-targeted attacks sometimes cast large but random changes in {\em logits}. These cases are not recoverable without extra information. 

\begin{table}[t]
\centering
\caption{Classification accuracy on clean and PGD-attacked images of different networks. Here, {\em full} indicates that the entire ILSVRC2012 validation set ($50\rm{,}000$ images) are used, while {\em selected} refers to a set of $1\rm{,}000$ images that are correctly classified by ResNet-50.}
\label{tab:network}
\resizebox{\textwidth}{!}{
\begin{tabular}{@{}ccccccccc@{}}
\toprule
             & \multicolumn{2}{c}{Clean, {\em full}} & \multicolumn{2}{c}{PGD, {\em full}} & \multicolumn{2}{c}{Clean, {\em selected}} & \multicolumn{2}{c}{PGD, {\em selected}} \\ \cmidrule(l){2-9}
             & No Defense   & Corrected     & No defense    & Corrected     & No defense    & Corrected     & No defense    & Corrected     \\ \midrule
VGG-16       & $73.48\%$    & $69.43\%$     & $0.04\%$      & $67.47\%$     & $94.7\%$      & $90.8\%$      & $0.0\%$       & $71.8\%$      \\
ResNet-50    & $76.13\%$    & $73.85\%$     & $0.00\%$      & $72.39\%$     & $100.0\%$     & $96.5\%$      & $0.0\%$       & $75.4\%$      \\
DenseNet-121 & $74.47\%$    & $72.20\%$     & $0.01\%$      & $88.29\%$     & $100.0\%$     & $95.3\%$      & $0.0\%$       & $90.5\%$      \\ \bottomrule
\end{tabular}}
\end{table}

Besides, a surprising result produced by our approach is that the classification accuracy on DenseNet-121, after being attacked and recovered, even improves from $74.47\%$ to $88.29\%$. This phenomenon is similar to `label leaking'~\citep{Kurakin2017AdversarialML}, which claimed that the accuracy on adversarial images gets much higher than that on clean images for a model adversarially trained on FGSM adversarial examples. However, label leaking was found to only occur with one-step attacks that use the true labels and vanish if an iterative method is used. In our experiment settings, PGD attack is used which is a very strong iterative method with randomness that further increases uncertainty. In addition, our correction network merely takes {\em logits} as input but cannot directly access the transformed image. This finding extends the problem of label leaking and reveals the possibility of helping classifiers with adversarial examples.

Next, we evaluate our approach on several state-of-the-art adversarial attacks, including PGD~\citep{Madry2018TowardsDL}, MIM~\citep{Dong2018BoostingAA}, DeepFool~\citep{MoosaviDezfooli2016DeepFoolAS} and C\&W~\citep{Carlini2017TowardsET}. A pre-trained ResNet-50 model is used as the target, and all training settings simply remain unchanged as in previous experiments. Differently, we only use $1$ test image per class ($1\rm{,}000$ in total) that can be correctly classified by ResNet-50 in the test stage, which is mainly due to the slowness of the DeepFool attack. Experimental results are summarized in Table~\ref{tab:attack}, from which one can observe quite similar phenomena as in the previous experiments. Here, we draw a few comments on the different properties among these attackers. For PGD, {\em logits} correction manages to recover $75.4\%$ of adversarial examples and still maintains a sufficiently high accuracy ($96.5\%$) on clean examples. MIM differs from PGD in that no random start is used and a momentum term is introduced to stabilize gradient updates, which results in less diversity and uncertainty. As a result, {\em logits} correction is able to learn better class-level relationship and thus recovers a larger fraction ($87.8\%$) of adversarial examples. As for DeepFool, note that image distortions are relatively smaller since it minimizes the perturbations in $\ell_2$-norm. Nevertheless, our method still succeeds in correcting $76.1\%$ of adversarial {\em logits}. Among all evaluated attacks, C\&W is the most difficult to defend, with our approach yielding an accuracy of less than $70\%$ on adversarial examples and the accuracy on clean examples is also largely affected. This is partly because C\&W, besides controlling the $\ell_2$-norm like DeepFool, uses a different kind of objective function and explicitly optimizes the magnitude of perturbations, which results in quite different behaviors in adversarial {\em logits} and thus increases the difficulty of correction.

\begin{table}[t]
\begin{minipage}[b]{0.54\textwidth}
\centering
\caption{Classification accuracy on clean and different adversarial images of ResNet-50. For fair comparison, we use the {\em selected} test set containing $1\rm{,}000$ images whose clean version is correctly recognized by ResNet-50.}
\label{tab:attack}
\resizebox{\linewidth}{!}{
\begin{tabular}{@{}ccccc@{}}
\toprule
         & \multicolumn{2}{c}{Clean, {\em selected}} & \multicolumn{2}{c}{Adversarial, {\em selected}} \\ \cmidrule(l){2-5}
         & No defense           & Corrected          & No defense           & Corrected                \\ \midrule
PGD      & $100.0\%$            & $96.5\%$           & $0.0\%$              & $75.4\%$                 \\
MIM      & $100.0\%$            & $96.3\%$           & $0.1\%$              & $87.8\%$                 \\
DeepFool & $100.0\%$            & $97.0\%$           & $0.0\%$              & $76.1\%$                 \\
C\&W     & $100.0\%$            & $82.8\%$           & $0.0\%$              & $66.8\%$                 \\ \bottomrule
\end{tabular}
}
\end{minipage}
\hfill
\begin{minipage}[b]{0.43\textwidth}
\centering
\caption{Classification accuracy when a defender trained on one attack is used to defend other attacks. The target model is ResNet-50, and all results are produced on the {\em selected} test set.}
\label{tab:transfer}
\resizebox{\linewidth}{!}{
\begin{tabular}{@{}ccccc@{}}
\toprule
         & \multicolumn{4}{c}{Defender Trained on}   \\ \cmidrule(l){2-5} 
         & PGD      & MIM      & DeepFool & C\&W     \\ \midrule
PGD      & $75.4\%$ & $70.3\%$ & $ 0.0\%$ & $ 4.4\%$ \\
MIM      & $85.7\%$ & $87.8\%$ & $ 0.1\%$ & $ 5.2\%$ \\
DeepFool & $54.5\%$ & $52.7\%$ & $76.1\%$ & $73.5\%$ \\
C\&W     & $10.3\%$ & $ 9.1\%$ & $19.2\%$ & $66.8\%$ \\ \bottomrule
\end{tabular}
}
\end{minipage}
\end{table}

Finally, we evaluate transferability, {\em i.e.}, whether a {\em logits} correction network trained on a specific attack can be used to defend other attacks on the same model. We fix ResNet-50 to be the target model, and evaluate the defender trained on each of the four attackers. Results are summarized in Table~\ref{tab:transfer}, in which each row corresponds to an attacker and each column a defender --- note that the diagonal is the same as the last column of Table~\ref{tab:attack}. One can see from the table that the defenders trained on PGD and MIM transfer well to each other, mainly due to the similar nature of these two attackers. These two defenders are also able to correct more than half of adversarial examples generated by DeepFool, which shows a wider aspect in generalization. On the contrary, the defender trained on DeepFool can hardly recover those adversarial examples produced by PGD and MIM, indicating that DeepFool, being an $\ell_2$-norm attacker, has different properties and thus the learned patterns for defense are less transferable. Similarly, C\&W has a closer behavior to DeepFool, an $\ell_2$-norm attacker, than to PGD and MIM, two $\ell_\infty$-norm attackers, which also reflects in the low recovery rates in the last row and column of Table~\ref{tab:transfer}. It is interesting to see a high transfer accuracy from C\&W to DeepFool, but a low accuracy in the opposite direction. This is because both DeepFool and C\&W are $\ell_2$-norm attackers, but the adversarial patterns generated by DeepFool are relatively simpler. So, the defender trained on C\&W can cover the patterns of DeepFool ($73.5\%$ of cases are defended), but the opposite is not true (only $19.2\%$ of cases are defended).

In the next section, we will provide a new insight to transferability, which focuses on finding the supporting classes of each defense and measuring the overlapping ratio between different sets of supporting classes.


\section{Explaining {\em logits}-based defense}
\label{explanation}


\subsection{How {\em logits} correction works}
\label{explanation:mechanism}

It remains an important topic to explain how our defender works. Thanks to the semantic basis and simplicity of our approach, for each defender, we can find a small set of classes that make most significant contributions to defense.

Given a clean vector of {\em logits}, $\mathbf{z}$, or an adversarial one, $\mathbf{z}^\ast$, the {\em logits} correction network $g(\cdot)$ acts as a multi-variate mapping between two $\mathbb{R}^C$ spaces where $C=1\rm{,}000$ in the ILSVRC2012 dataset~\citep{Russakovsky2015ImageNetLS}. Let $z_k$ denote the score of the $k$-th class in $\mathbf{z}$, and $g(\mathbf{z})_i$ the score of the $i$-th class in $g(\mathbf{z})$. The core idea is to compute the partial derivative $H_{i,k}(\mathbf{z})\doteq\frac{\partial g(\mathbf{z})_i}{\partial z_k}$, so that we can estimate the contribution of each element in the input {\em logits} to the corrected {\em logits}.

Suppose we have an adversarial sample $\mathbf{z}^\ast$ that has a ground-truth label of $i$ but is misclassified as class $j$ after being attacked. When feeding it into the correction network $g(\cdot)$, it should be recovered and thus $g(\mathbf{z}^\ast)_i$ should have the greatest score. To find out how $g(\cdot)$ manages to perform correction, we can either investigate how $g(\mathbf{z}^\ast)_i$ is `pulled up' by finding the classes that have high positive $H_{i,k}(\mathbf{z}^\ast)$ values, or how other classes are `pushed down' by finding the classes that have high negative impacts on the average of all {\em logits}, namely, $H_{\mathrm{mean},k}(\mathbf{z}^\ast)=\frac{1}{C}{\sum_l}H_{l,k}(\mathbf{z}^\ast)$.

We first explore an example using the {\em logits} correction network trained to defend the PGD attack~\citep{Madry2018TowardsDL} on ResNet-50~\citep{He2016DeepRL}. For each adversarial input $\mathbf{z}^\ast$ in the validation set, we find out $10$ greatest entries of $H_{i,k}(\mathbf{z}^\ast)$, with $i$ being the original label and $k$ ranges among all $1\rm{,}000$ classes. Interestingly, for a large amount of cases, no matter what the ground-truth label or the input image is, there always exist some specific classes that contribute most to recovering the correct label. We count over all $50\rm{,}000$ validation images, and find out that the greatest $H_{i,k}(\mathbf{z}^\ast)$ appears mostly when $k$ equals to $705$, $830$, $600$, $356$, $850$ and $447$. When we compute $-H_{\mathrm{mean},k}(\mathbf{z}^\ast)$ instead, most cases have the highest response in the $640$-th class, ({\em i.e.}, {\em `manhole cover'}). Similar phenomena are also found when we use PGD to attack other target networks, including VGG-16~\citep{Simonyan2015VeryDC} and DenseNet-121~\citep{Huang2017DenselyCC}. Some of the classes that contribute most to $H_{i,k}(\mathbf{z}^\ast)$ overlap with those for ResNet-50, implying that these classes are fragile to PGD. When we compute $-H_{\mathrm{mean},k}(\mathbf{z}^\ast)$ instead, the $640$-th class still dominants for both DenseNet-121 and VGG-16.


\subsection{Supporting classes and their relationship to transferability of defense}
\label{explanation:transferability}

Here, we define a new concept named {\bf supporting classes} as those classes that contribute most to {\em logits}-based defense. Taking both positive (`pulling up') and negative (`pushing down') effects into consideration, we compute $S_k\doteq H_{i,k}(\mathbf{z}^\ast)-H_{\mathrm{mean},k}(\mathbf{z}^\ast)$. Among all classes, $10$ classes of $k$ with greatest $S_k$ values are taken out for each image. We count the occurrences over the {\em selected} test set ($1\rm{,}000$ images), and finally pick up $10$ classes that appear most frequently as the supporting classes for defending a specific attack from the model. The supporting classes of defending the PGD attack on ResNet-50 are illustrated in Figure~\ref{fig:supporting_classes}, and we also show other cases in the supplementary material. We first emphasize that, indeed, these classes are the key to defense. On an arbitrary image, if we reduce the {\em logit} values of these supporting classes by $20$ and feed the modified {\em logits} into the trained defender, it is almost for sure that {\em logits} correction fails dramatically and, very likely, the classification result remains to be $j$, the originally misclassified class, even when the original or adversarial classes of this case are almost irrelevant to these supporting classes.

\begin{wrapfigure}{r}{7cm}
\centering
\includegraphics[width=\linewidth]{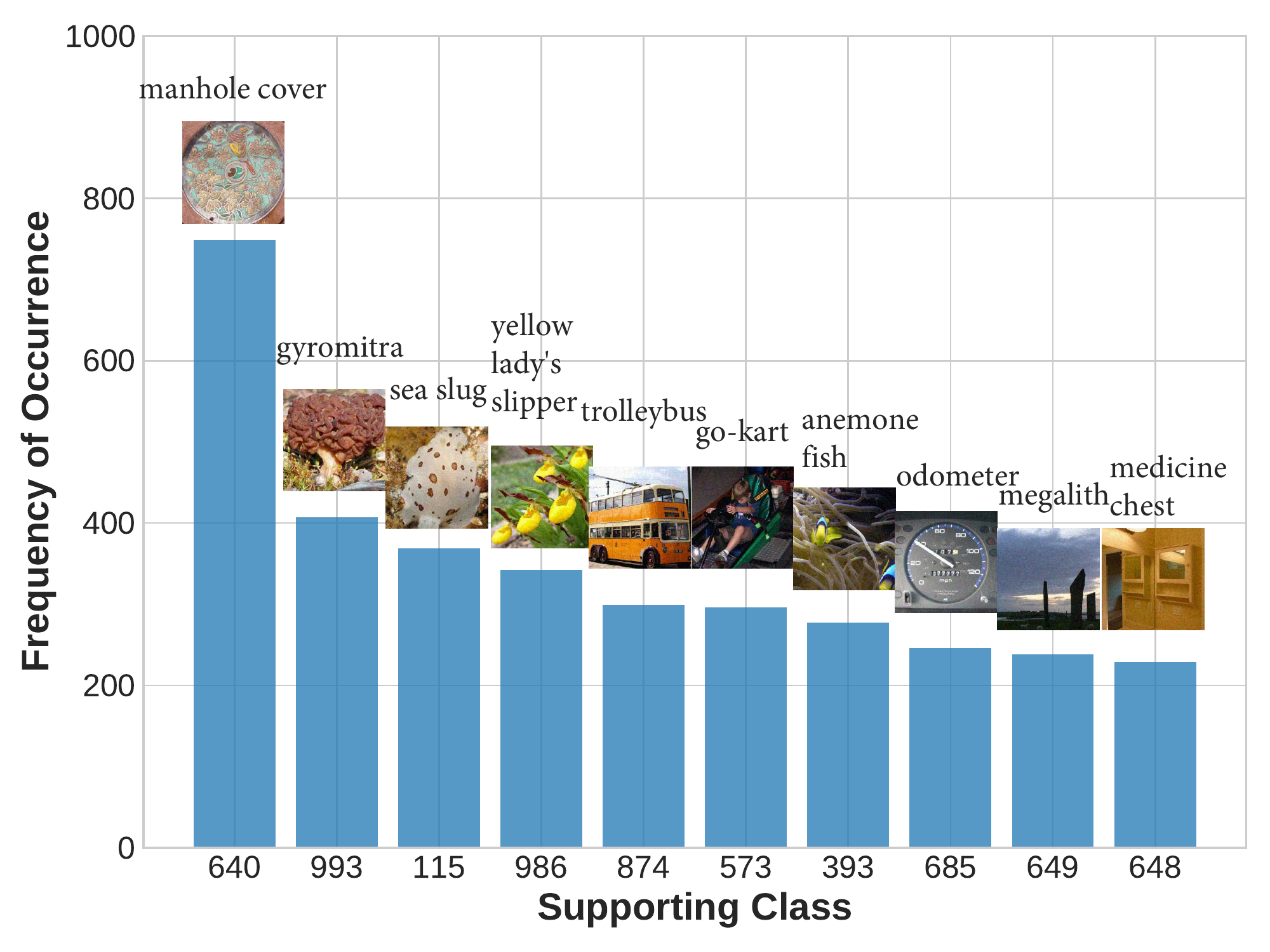}
\caption{Supporting classes of the PGD defender on ResNet-50. We list $10$ classes that appear most frequently in the top-10 of $S_k$, with the frequency of occurrence recorded on the vertical axis. For better visualization, we list the name of each class and attach a representative image above the bar.}
\label{fig:supporting_classes}
\end{wrapfigure}

As a further analysis, we reveal the relationship between the overlapping ratio of supporting classes and the transferability of our defender from one setting to another. We compute the Bhattacharyya coefficients between the sets of supporting classes produced by the four attacks. The first pair is PGD and MIM, which reports a high coefficient of $0.9662$. This implies that PGD and MIM have quite similar behavior in attacks, which is mainly due to their similar mechanisms ({\em e.g.}, $\ell_\infty$-bounded, iteration-based, {\em etc.}). Consequently, as shown in Table~\ref{tab:transfer}, the defenders trained on PGD and MIM transfer well to each other. A similar phenomenon appears in the pair of DeepFool and C\&W, two $\ell_2$-norm attackers, with an coefficient of $0.9222$. As a result, the defenders trained on DeepFool and C\&W produce the best transfer accuracy on each other (as we explained before, the weak transferability of the DeepFool defender is mainly because DeepFool is very easy to defend). Across $\ell_\infty$-norm and $\ell_2$-norm attackers, we report coefficients of $0.9004$ for the pair of PGD and DeepFool, and $0.8623$ for PGD and C\&W, respectively. Note that both numbers are less than those between the same type of attacks. In addition, the defender trained on PGD achieves an accuracy of $54.5\%$ on DeepFool, and merely $10.3\%$ on C\&W, which aligns with the coefficient values. {\bf These results verify our motivation --- {\em logits}-based correction is easier to be explained, in particular at the semantic level.}

Last but not least, the impact of supporting classes can also be analyzed in the instance level, {\em i.e.}, finding the classes that contribute most to defending each attack on each single image. We observe a few interesting phenomena, {\em e.g.}, for PGD and MIM, the top-1 supporting class is very likely to be the $640$-th class, while for DeepFool, the most important class is always the ground-truth class which differs from case to case. This partly explains the transferability between PGD/MIM and DeepFool. More instance-level analysis is provided in the supplementary material.


\section{Conclusions}
\label{conclusions}

In this paper, we find that a wide range of state-of-the-art adversarial attacks can be defended by merely correcting {\em logits}, the features produced by the last layer of a deep neural network. This implies that each attacker leaves `fingerprints' during the attack. Although it is difficult to make rules to detect and eliminate such impacts, we design a learning-based approach, simple but effective, which achieves high recovery rates in a few combinations of attacks and target networks. Going one step forward, we reveal that our defender works by finding a few supporting classes for each attack-network combination, and by checking the overlapping ratio of these classes, we can estimate the transferability of a defense across different scenarios.

Our research leaves a few unsolved problems. For example, it is unclear whether there exists an attack algorithm that cannot be corrected by our defender, or if we can find deeper connections between our discovery and the mechanism of deep neural networks. In addition, we believe that improving the transferability of this defender is a promising direction, in which we shall continue in the future.

{\small
\bibliography{ref}

\begin{thebibliography}{36}
\providecommand{\natexlab}[1]{#1}
\providecommand{\url}[1]{\texttt{#1}}
\expandafter\ifx\csname urlstyle\endcsname\relax
  \providecommand{\doi}[1]{doi: #1}\else
  \providecommand{\doi}{doi: \begingroup \urlstyle{rm}\Url}\fi

\bibitem[Angluin and Laird(1988)]{Angluin1987LearningFN}
Dana Angluin and Philip~D. Laird.
\newblock Learning from noisy examples.
\newblock \emph{Machine Learning}, 2:\penalty0 343--370, 1988.

\bibitem[Athalye et~al.(2018)Athalye, Carlini, and
  Wagner]{Athalye2018ObfuscatedGG}
Anish Athalye, Nicholas Carlini, and David~A. Wagner.
\newblock Obfuscated gradients give a false sense of security: Circumventing
  defenses to adversarial examples.
\newblock In \emph{ICML}, 2018.

\bibitem[Buckman et~al.(2018)Buckman, Roy, Raffel, and
  Goodfellow]{Buckman2018ThermometerEO}
Jacob Buckman, Aurko Roy, Colin~A. Raffel, and Ian~J. Goodfellow.
\newblock Thermometer encoding: One hot way to resist adversarial examples.
\newblock In \emph{ICLR}, 2018.

\bibitem[Carlini and Wagner(2017)]{Carlini2017TowardsET}
Nicholas Carlini and David~A. Wagner.
\newblock Towards evaluating the robustness of neural networks.
\newblock In \emph{IEEE Symposium on Security and Privacy (SP)}, 2017.

\bibitem[Dong et~al.(2018)Dong, Liao, Pang, Su, Zhu, Hu, and
  Li]{Dong2018BoostingAA}
Yinpeng Dong, Fangzhou Liao, Tianyu Pang, Hang Su, Jun Zhu, Xiaolin Hu, and
  Jianguo Li.
\newblock Boosting adversarial attacks with momentum.
\newblock In \emph{CVPR}, 2018.

\bibitem[Dziugaite et~al.(2016)Dziugaite, Ghahramani, and
  Roy]{Dziugaite2016ASO}
Gintare~Karolina Dziugaite, Zoubin Ghahramani, and Daniel~M. Roy.
\newblock A study of the effect of jpg compression on adversarial images.
\newblock \emph{CoRR}, abs/1608.00853, 2016.

\bibitem[Goldberger and Ben-Reuven(2017)]{Goldberger2017TrainingDN}
Jacob Goldberger and Ehud Ben-Reuven.
\newblock Training deep neural-networks using a noise adaptation layer.
\newblock In \emph{ICLR}, 2017.

\bibitem[Goodfellow et~al.(2015)Goodfellow, Shlens, and
  Szegedy]{Goodfellow2015ExplainingAH}
Ian~J. Goodfellow, Jonathon Shlens, and Christian Szegedy.
\newblock Explaining and harnessing adversarial examples.
\newblock In \emph{ICLR}, 2015.

\bibitem[Guo et~al.(2018)Guo, Rana, Ciss{\'e}, and van~der
  Maaten]{Guo2018CounteringAI}
Chuan Guo, Mayank Rana, Moustapha Ciss{\'e}, and Laurens van~der Maaten.
\newblock Countering adversarial images using input transformations.
\newblock In \emph{ICLR}, 2018.

\bibitem[He et~al.(2016)He, Zhang, Ren, and Sun]{He2016DeepRL}
Kaiming He, Xiangyu Zhang, Shaoqing Ren, and Jian Sun.
\newblock Deep residual learning for image recognition.
\newblock In \emph{CVPR}, 2016.

\bibitem[Hinton et~al.(2012)Hinton, Srivastava, Krizhevsky, Sutskever, and
  Salakhutdinov]{Hinton2012ImprovingNN}
Geoffrey~E. Hinton, Nitish Srivastava, Alex Krizhevsky, Ilya Sutskever, and
  Ruslan~R. Salakhutdinov.
\newblock Improving neural networks by preventing co-adaptation of feature
  detectors.
\newblock \emph{CoRR}, abs/1207.0580, 2012.

\bibitem[Huang et~al.(2017)Huang, Liu, and Weinberger]{Huang2017DenselyCC}
Gao Huang, Zhuang Liu, and Kilian~Q. Weinberger.
\newblock Densely connected convolutional networks.
\newblock In \emph{CVPR}, 2017.

\bibitem[Kannan et~al.(2018)Kannan, Kurakin, and
  Goodfellow]{Kannan2018AdversarialLP}
Harini Kannan, Alexey Kurakin, and Ian~J. Goodfellow.
\newblock Adversarial logit pairing.
\newblock In \emph{NeurIPS}, 2018.

\bibitem[Kingma and Ba(2015)]{Kingma2015AdamAM}
Diederik~P. Kingma and Jimmy Ba.
\newblock Adam: A method for stochastic optimization.
\newblock In \emph{ICLR}, 2015.

\bibitem[Krizhevsky et~al.(2012)Krizhevsky, Sutskever, and
  Hinton]{Krizhevsky2012ImageNetCW}
Alex Krizhevsky, Ilya Sutskever, and Geoffrey~E. Hinton.
\newblock Imagenet classification with deep convolutional neural networks.
\newblock In \emph{NeurIPS}, 2012.

\bibitem[Kurakin et~al.(2017{\natexlab{a}})Kurakin, Goodfellow, and
  Bengio]{Kurakin2017AdversarialEI}
Alexey Kurakin, Ian~J. Goodfellow, and Samy Bengio.
\newblock Adversarial examples in the physical world.
\newblock In \emph{ICLR Workshop}, 2017{\natexlab{a}}.

\bibitem[Kurakin et~al.(2017{\natexlab{b}})Kurakin, Goodfellow, and
  Bengio]{Kurakin2017AdversarialML}
Alexey Kurakin, Ian~J. Goodfellow, and Samy Bengio.
\newblock Adversarial machine learning at scale.
\newblock In \emph{ICLR}, 2017{\natexlab{b}}.

\bibitem[Liao et~al.(2018)Liao, Liang, Dong, Pang, Zhu, and
  Hu]{Liao2018DefenseAA}
Fangzhou Liao, Ming Liang, Yinpeng Dong, Tianyu Pang, Jun Zhu, and Xiaolin Hu.
\newblock Defense against adversarial attacks using high-level representation
  guided denoiser.
\newblock In \emph{CVPR}, 2018.

\bibitem[Madry et~al.(2018)Madry, Makelov, Schmidt, Tsipras, and
  Vladu]{Madry2018TowardsDL}
Aleksander Madry, Aleksandar Makelov, Ludwig Schmidt, Dimitris Tsipras, and
  Adrian Vladu.
\newblock Towards deep learning models resistant to adversarial attacks.
\newblock In \emph{ICLR}, 2018.

\bibitem[Moosavi-Dezfooli et~al.(2016)Moosavi-Dezfooli, Fawzi, and
  Frossard]{MoosaviDezfooli2016DeepFoolAS}
Seyed-Mohsen Moosavi-Dezfooli, Alhussein Fawzi, and Pascal Frossard.
\newblock Deepfool: A simple and accurate method to fool deep neural networks.
\newblock In \emph{CVPR}, 2016.

\bibitem[Nayebi and Ganguli(2017)]{Nayebi2017BiologicallyIP}
Aran Nayebi and Surya Ganguli.
\newblock Biologically inspired protection of deep networks from adversarial
  attacks.
\newblock \emph{CoRR}, abs/1703.09202, 2017.

\bibitem[Osadchy et~al.(2017)Osadchy, Hernandez-Castro, Gibson, Dunkelman, and
  P{\'e}rez-Cabo]{Osadchy2017NoBE}
Margarita Osadchy, Julio Hernandez-Castro, Stuart~J. Gibson, Orr Dunkelman, and
  Daniel P{\'e}rez-Cabo.
\newblock No bot expects the deepcaptcha! introducing immutable adversarial
  examples, with applications to captcha generation.
\newblock In \emph{IEEE Transactions on Information Forensics and Security},
  volume~12, pages 2640--2653, 2017.

\bibitem[Papernot et~al.(2016{\natexlab{a}})Papernot, Faghri, Carlini,
  Goodfellow, Feinman, Kurakin, Xie, Sharma, Brown, Roy, Matyasko, Behzadan,
  Hambardzumyan, Zhang, Juang, Li, Sheatsley, Garg, Uesato, Gierke, Dong,
  Berthelot, Hendricks, Rauber, Long, and McDaniel]{papernot2016TechnicalRO}
Nicolas Papernot, Fartash Faghri, Nicholas Carlini, Ian Goodfellow, Reuben
  Feinman, Alexey Kurakin, Cihang Xie, Yash Sharma, Tom Brown, Aurko Roy,
  Alexander Matyasko, Vahid Behzadan, Karen Hambardzumyan, Zhishuai Zhang,
  Yi-Lin Juang, Zhi Li, Ryan Sheatsley, Abhibhav Garg, Jonathan Uesato, Willi
  Gierke, Yinpeng Dong, David Berthelot, Paul Hendricks, Jonas Rauber, Rujun
  Long, and Patrick McDaniel.
\newblock Technical report on the cleverhans v2.1.0 adversarial examples
  library.
\newblock \emph{CoRR}, abs/1610.00768, 2016{\natexlab{a}}.

\bibitem[Papernot et~al.(2016{\natexlab{b}})Papernot, McDaniel, Wu, Jha, and
  Swami]{Papernot2016DistillationAA}
Nicolas Papernot, Patrick~D. McDaniel, Xi~Wu, Somesh Jha, and Ananthram Swami.
\newblock Distillation as a defense to adversarial perturbations against deep
  neural networks.
\newblock In \emph{IEEE Symposium on Security and Privacy (SP)},
  2016{\natexlab{b}}.

\bibitem[Papernot et~al.(2017)Papernot, McDaniel, Goodfellow, Jha, Celik, and
  Swami]{Papernot2017PracticalBA}
Nicolas Papernot, Patrick~D. McDaniel, Ian~J. Goodfellow, Somesh Jha, Z.~Berkay
  Celik, and Ananthram Swami.
\newblock Practical black-box attacks against machine learning.
\newblock In \emph{Proceedings of the 2017 ACM on Asia Conference on Computer
  and Communications Security}, 2017.

\bibitem[Paszke et~al.(2017)Paszke, Gross, Chintala, Chanan, Yang, DeVito, Lin,
  Desmaison, Antiga, and Lerer]{Paszke2017AutomaticDI}
Adam Paszke, Sam Gross, Soumith Chintala, Gregory Chanan, Edward Yang, Zachary
  DeVito, Zeming Lin, Alban Desmaison, Luca Antiga, and Adam Lerer.
\newblock Automatic differentiation in pytorch.
\newblock In \emph{NeurIPS Workshop}, 2017.

\bibitem[Patrini et~al.(2017)Patrini, Rozza, Menon, Nock, and
  Qu]{Patrini2017MakingDN}
Giorgio Patrini, Alessandro Rozza, Aditya~Krishna Menon, Richard Nock, and
  Lizhen Qu.
\newblock Making deep neural networks robust to label noise: A loss correction
  approach.
\newblock In \emph{CVPR}, 2017.

\bibitem[Prakash et~al.(2018)Prakash, Moran, Garber, DiLillo, and
  Storer]{Prakash2018DeflectingAA}
Aaditya Prakash, Nick Moran, Solomon Garber, Antonella DiLillo, and James~A.
  Storer.
\newblock Deflecting adversarial attacks with pixel deflection.
\newblock In \emph{CVPR}, 2018.

\bibitem[Roth et~al.(2019)Roth, Kilcher, and Hofmann]{Roth2019TheOA}
Kevin Roth, Yannic Kilcher, and Thomas Hofmann.
\newblock The odds are odd: A statistical test for detecting adversarial
  examples.
\newblock \emph{CoRR}, abs/1902.04818, 2019.

\bibitem[Russakovsky et~al.(2015)Russakovsky, Deng, Su, Krause, Satheesh, Ma,
  Huang, Karpathy, Khosla, Bernstein, Berg, and
  Fei-Fei]{Russakovsky2015ImageNetLS}
Olga Russakovsky, Jia Deng, Hao Su, Jonathan Krause, Sanjeev Satheesh, Sean Ma,
  Zhiheng Huang, Andrej Karpathy, Aditya Khosla, Michael~S. Bernstein,
  Alexander~C. Berg, and Li~Fei-Fei.
\newblock Imagenet large scale visual recognition challenge.
\newblock In \emph{IJCV}, volume 115, pages 211--252, 2015.

\bibitem[Simonyan and Zisserman(2015)]{Simonyan2015VeryDC}
Karen Simonyan and Andrew Zisserman.
\newblock Very deep convolutional networks for large-scale image recognition.
\newblock In \emph{ICLR}, 2015.

\bibitem[Szegedy et~al.(2014)Szegedy, Zaremba, Sutskever, Bruna, Erhan,
  Goodfellow, and Fergus]{Szegedy2014IntriguingPO}
Christian Szegedy, Wojciech Zaremba, Ilya Sutskever, Joan Bruna, Dumitru Erhan,
  Ian~J. Goodfellow, and Rob Fergus.
\newblock Intriguing properties of neural networks.
\newblock In \emph{ICLR}, 2014.

\bibitem[Tram{\`e}r et~al.(2018)Tram{\`e}r, Kurakin, Papernot, Goodfellow,
  Boneh, and McDaniel]{Tramr2018EnsembleAT}
Florian Tram{\`e}r, Alexey Kurakin, Nicolas Papernot, Ian~J. Goodfellow, Dan
  Boneh, and Patrick~D. McDaniel.
\newblock Ensemble adversarial training: Attacks and defenses.
\newblock In \emph{ICLR}, 2018.

\bibitem[Xie et~al.(2017)Xie, Wang, Zhang, Zhou, Xie, and
  Yuille]{Xie2017AdversarialEF}
Cihang Xie, Jianyu Wang, Zhishuai Zhang, Yuyin Zhou, Lingxi Xie, and Alan~L.
  Yuille.
\newblock Adversarial examples for semantic segmentation and object detection.
\newblock In \emph{ICCV}, 2017.

\bibitem[Xie et~al.(2018{\natexlab{a}})Xie, Wang, Zhang, Ren, and
  Yuille]{Xie2018MitigatingAE}
Cihang Xie, Jianyu Wang, Zhishuai Zhang, Zhou Ren, and Alan~Loddon Yuille.
\newblock Mitigating adversarial effects through randomization.
\newblock In \emph{ICLR}, 2018{\natexlab{a}}.

\bibitem[Xie et~al.(2018{\natexlab{b}})Xie, Wu, van~der Maaten, Yuille, and
  He]{Xie2018FeatureDF}
Cihang Xie, Yuxin Wu, Laurens van~der Maaten, Alan~Loddon Yuille, and Kaiming
  He.
\newblock Feature denoising for improving adversarial robustness.
\newblock \emph{CoRR}, abs/1812.03411, 2018{\natexlab{b}}.

\end{thebibliography}
}

\clearpage
\appendix

\section{Supporting classes of different attacks}

\begin{figure}[htb]
\centering
\includegraphics[width=\textwidth]{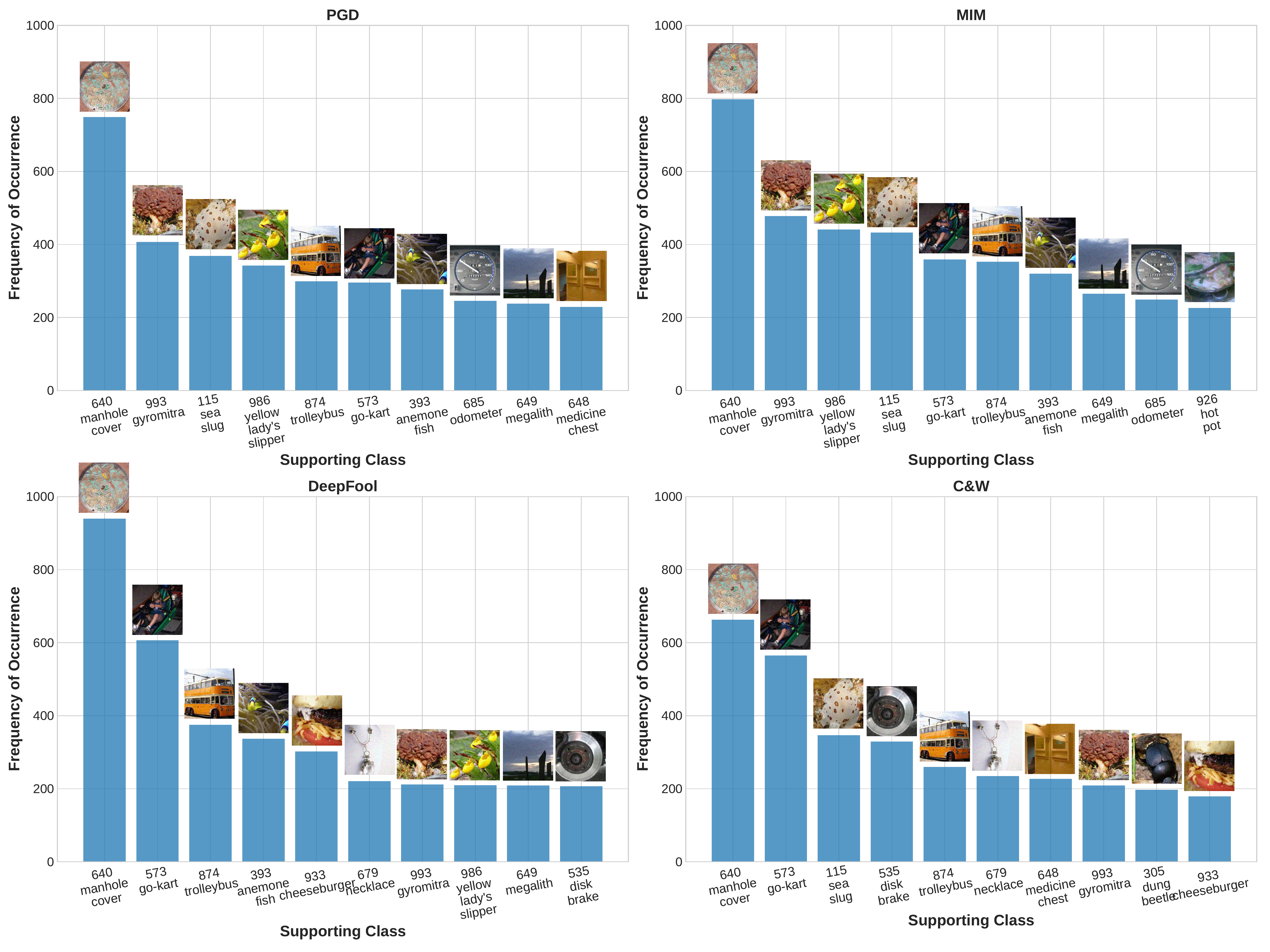}
\caption{Supporting classes of each adversarial attack on ResNet-50. We list $10$ classes that appear most frequently in the top-$10$ of $S_k$ (see Section \ref{explanation:transferability}), with the frequency of occurrences recorded on the vertical axis. For better visualization, we list the name of each class on the horizontal axis, and also attach a representative image above the bar. Please zoom in for better clarity.}
\label{fig:supporting_classes_supp}
\end{figure}

In Figure~\ref{fig:supporting_classes_supp}, we illustrate the supporting classes of defending PGD~\citep{Madry2018TowardsDL}, MIM~\citep{Dong2018BoostingAA}, DeepFool~\citep{MoosaviDezfooli2016DeepFoolAS} and C\&W~\citep{Carlini2017TowardsET} on ResNet-50~\citep{He2016DeepRL}, respectively. Just like the cases of PGD attack on different target networks, these different attacks also share some supporting classes in common. Note that the supporting classes of PGD and MIM are nearly the same, even considering their relative order in frequency. This aligns with the high Bhattacharyya coefficients between them and the good transferability of defenders trained on them. The two $\ell_2$-norm attacks, DeepFool and C\&W, also have very similar supporting classes, and this similarity yet accounts for the transferability of their corresponding defenders. Besides, one can observe from Figure~\ref{fig:supporting_classes_supp} that MIM and DeepFool have stronger response in the $640$-th class, {\em i.e.}, {\em `manhole cover'}, which may explain why they are easier to be defended. Similarly, the difficulty of defending against C\&W may also reside in the fact that the supporting classes of this attack are not as strong as other attackers, {\em e.g.}, the dominant class is also the $640$-th class, but its frequency is relatively lower (also closer to the second class).

\section{Delving into supporting classes at instance level}

\begin{figure}[htb]
\centering
\includegraphics[width=\textwidth]{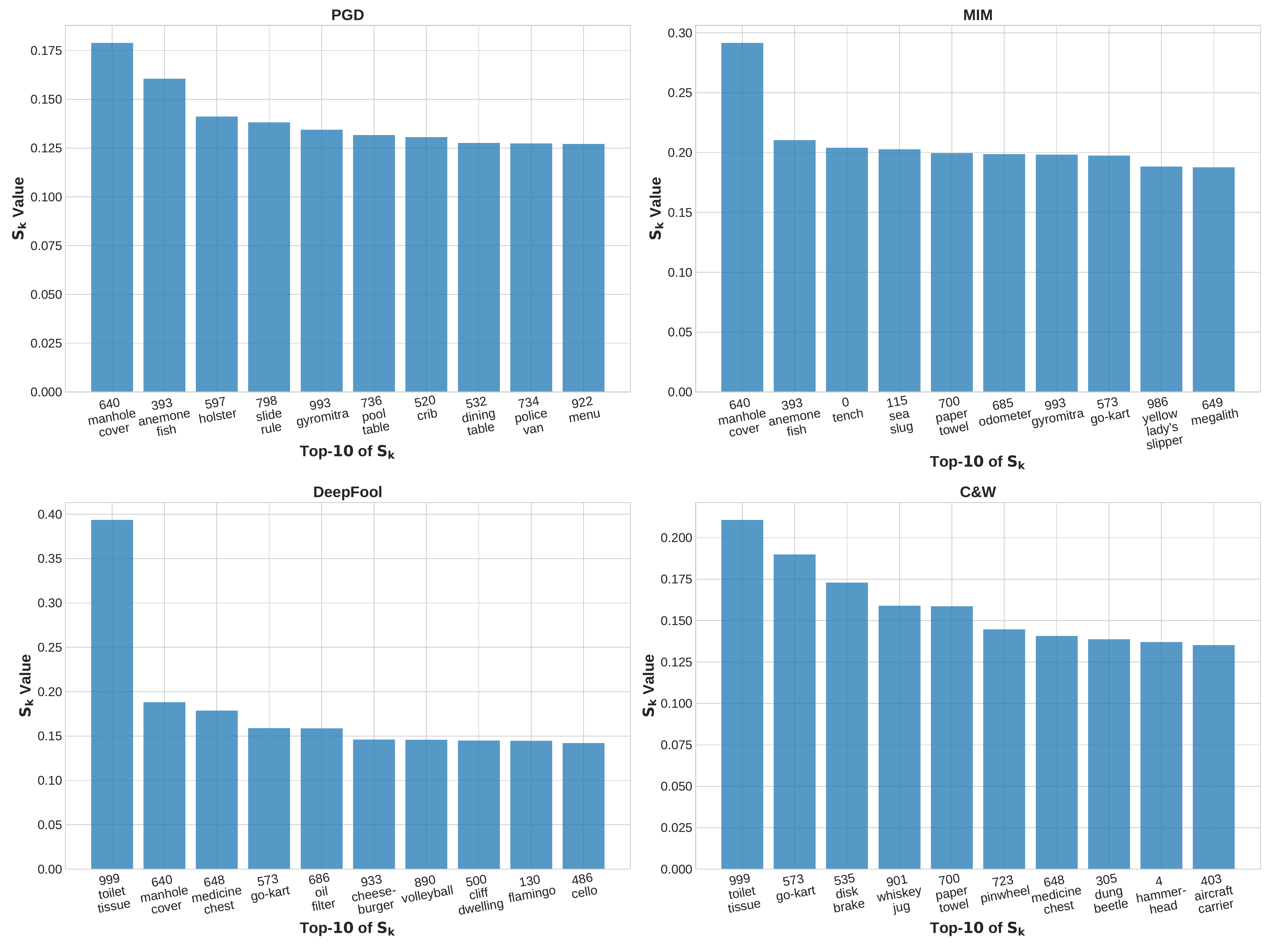}
\caption{Top-$10$ classes of $S_k$ and their corresponding values for an example with ground-truth label $999$ and attacked by PGD, MIM, DeepFool and C\&W, respectively. For better visualization, we list the name of each class. Please zoom in for better clarity.}
\label{fig:instance_level}
\end{figure}

To better understand the impact of supporting classes, we further inspect the classes that contribute most to defending a single example, {\em i.e.}, at an instance level. The top-$10$ classes of $S_k$ and their corresponding values are taken out for each image. Figure~\ref{fig:instance_level} shows such classes and values for an example with ground-truth label $999$ and attacked by the four attackers.

We first explore the PGD attacker~\citep{Madry2018TowardsDL} as usual, and find that while the supporting classes we obtain in the last part frequently appear in the top-$10$, the $640$-th class always occupies the top-$3$ and even top-$1$, especially when the attacked example is successfully corrected by the defender. This once again shows the importance of the $640$-th class, and similar phenomenon is found when the MIM attacker~\citep{Dong2018BoostingAA} is used.

As for DeepFool~\citep{MoosaviDezfooli2016DeepFoolAS}, things become different, as the top-$1$ of $S_k$ always lies in $i$, the ground-truth label, with a much greater value than the second most significant supporting class. This is mainly due to the design nature of DeepFool, which moves an example across the nearest decision boundary and thus the original class should still have a high score in the adversarial {\em logits} $\mathbf{z}^\ast$. In other words, the adversary of DeepFool can be recovered by assigning a greater weight to $i$. Consequently, although DeepFool shares a similar property that the $640$-th class still appears most frequently, it shows quite a different behavior in defense, which partly reflects in the transfer experiments between DeepFool and PGD. Given an example attacked by DeepFool and a defender trained on PGD, the defender can correct the example basing on the supporting classes rather than $i$, yielding a relatively good performance; On the contrary, given an example attacked by PGD and a defender trained on DeepFool, the defender will focus too much on the ground-truth class $i$ of the example, and thus fail to correct the attack of PGD which does not have such preference.

Finally, we study the case of C\&W~\cite{Carlini2017TowardsET}. We find that the set of supporting classes of $S_k$ lies between PGD and DeepFool, and is more similar to that of DeepFool ({\em i.e.}, the most significant class is usually the ground-truth class $i$, but sometimes $640$). This corresponds to the fact that the defender trained on C\&W can transfer to DeepFool better than PGD. However, since the behaviour of C\&W is more irregular ({\em e.g.}, $i$ is not always the most important contributor), it is more difficult to defend by defenders trained on other attacks.


\end{document}